# Underwater Optical Image Processing: A Comprehensive Review

Huimin Lu[1,*], Yujie Li[2,*], Yudong Zhang[3], Min Chen[4], Seiichi Serikawa[1], Hyoungseop Kim[1]

[1]Kyushu Institute of Technology, 1-1 Sensui, Tobata, Kitakyushu 804-8550, Japan
[2]Chinese Academy of Science, 7 Nanhai Rd., Shinan, Qingdao 266000, China
[3]Yangzhou University, 198 Huayang West Rd., Hanjiang, Yangzhou 225127, China
[4]Nanjing Normal University, 122 Ninghai Rd., Gulou, Nanjing 210097, China
[5]Huazhong University of Science and Technology, 1037 Luoyu Rd., Hongshan, Wuhan 430074, China

Corresponding author: Y. Li (yzyjli@gmail.com); H. Lu (dr.huimin.lu@ieee.org)

**Abstract:**
Underwater cameras are widely used to observe the sea floor. They are usually included in autonomous underwater vehicles (AUVs), unmanned underwater vehicles (UUVs), and in situ ocean sensor networks. Despite being an important sensor for monitoring underwater scenes, there exist many issues with recent underwater camera sensors. Because of light's transportation characteristics in water and the biological activity at the sea floor, the acquired underwater images often suffer from scatters and large amounts of noise. Over the last five years, many methods have been proposed to overcome traditional underwater imaging problems. This paper aims to review the state-of-the-art techniques in underwater image processing by highlighting the contributions and challenges presented in over 40 papers. We present an overview of various underwater image-processing approaches, such as underwater image de-scattering, underwater image color restoration, and underwater image quality assessments. Finally, we summarize the future trends and challenges in designing and processing underwater imaging sensors.

## 1. Introduction

Underwater optical imaging (OPI) is a challenging field in computer vision research [1]. As opposed to land photography, there exist many constraints in underwater imaging [2,3]. First, due to the medium, scattering always causes a blurring effect in underwater photography; this rarely occurs in land photography. Second, wavelength absorption usually causes a color reduction in the captured images, which rarely occurs in air. Third, except for electronic noise, the sediments in the water also affect high dimensional imaging. Another problem occurs because artificial lighting is widely used for underwater photography, and this non-uniform lighting causes vignetting in captured images [4]. Furthermore, the flickering affects always exist in sunshine day. This will cause the captured images with strong highlights in the shallow ocean. Consequently, underwater images have specific characteristics that need to be taken into account during gathering and processing. Figure 1 shows the concept map of ocean observing network, which is proposed by Lu et al. [2]. Cameras are usually used in many devices. Common issues with underwater images, such as light attenuation, scattering, non-uniform lighting, shadows, color shading, suspended particles, or the abundance of marine life, can be overcome via underwater optical image processing.

The volume scattering function describes the angular distribution of light scattered by the suspension of particles in a direction at a given wavelength. Scatters redirect the angle of the photon path, absorption removes the photons from the light path. Absorption in pure water indicates that blue wavelengths are more sensitive to absorption than red wavelengths. However, in phytoplankton water, red wavelengths are not terminated more than blue wavelengths. Therefore, it is difficult to measure absorption rates in practice. On the other hand, the wavelength absorption is relayed on the geographic location of the seawater. Different salinity of seawater has a different wavelength absorption coefficients.

In this paper, we first review two categories of underwater image de-scattering methods: hardware-based approaches and software-based approaches. Then, we summarize four typical underwater image color restoration methods. Next, we cover two underwater image quality assessment methods: reference-based indexes and non-reference indexes. Finally, we summarize this paper and elaborate upon future trends in this research field.

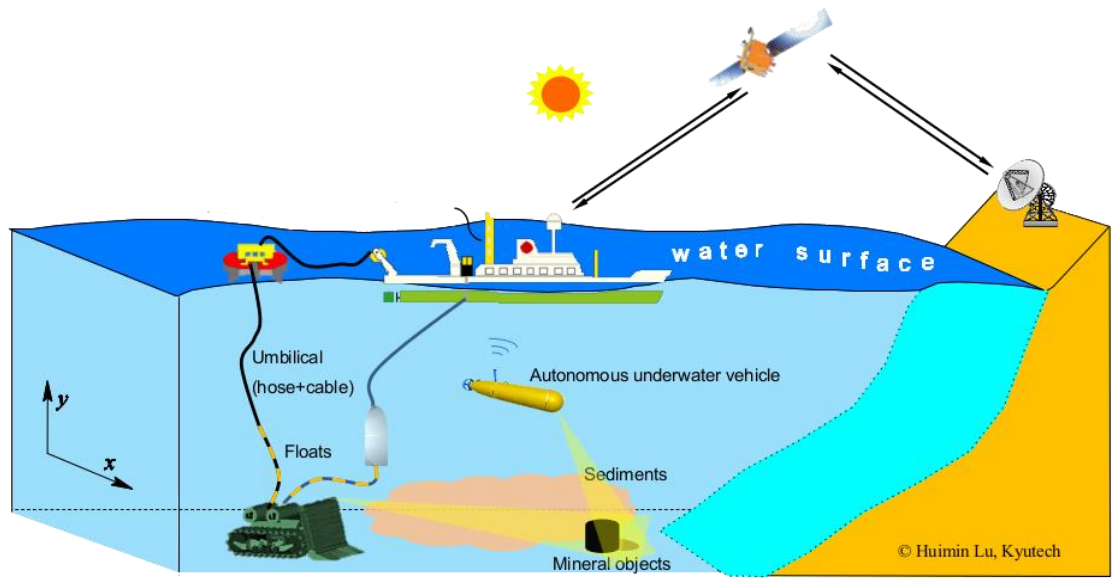

**Figure 1.** Concept Map of Ocean Observing Network

## 2. Types of Underwater Imaging

There are several existing methods for underwater image processing. In general, these methods can be categorized into two approaches: hardware-based methods [5,6,7,8,9,10] and software-based methods [11,12,13,14]. There are four traditional hardware approaches to underwater imaging: polarization, range-gated imaging, fluorescence imaging [32,33], and stereo imaging [34]. Light has the properties of intensity, wavelength, and polarization. Natural light is without polarization, while light reaching an imaging sensor often has biased polarization. Preliminary studies have verified that backscatter can be reduced using polarization. There are two classic methods for underwater polarization imaging, one is to use a polarization filter attached in front of the camera to receive the biased images [5,6]. The other method is to use a polarized light source to capture different illuminated images of the same scene. The polarization method is designed to capture images quickly as well as to reduce the noise significantly. Schecher et al. [6] proposed the state-of-the-art polarization imaging method for underwater. The result is shown in Figure 2.

Polarization imaging is a passive imaging method, while range-gated imaging is an active imaging method and widely used for laser-imaging systems in turbid water. The most of recent underwater laser-imaging methods were summarized in Ref. [8]. In a laser-imaging system, the camera is adjacent to the light source, and the target is behind the turbid medium. This system operates by selecting the reflected light from the object and blocking the backscatter by closing a flash gate. However, the laser-imaging methods have the disadvantage of being susceptible to environment and device setting are complex. Consequently, laser imaging instruments are seldom used in industrial applications.

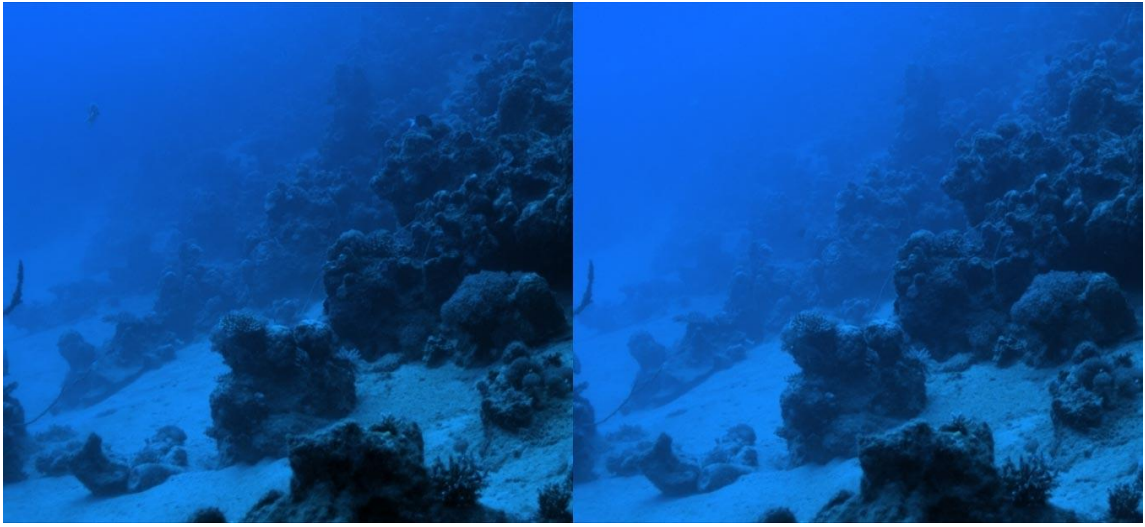

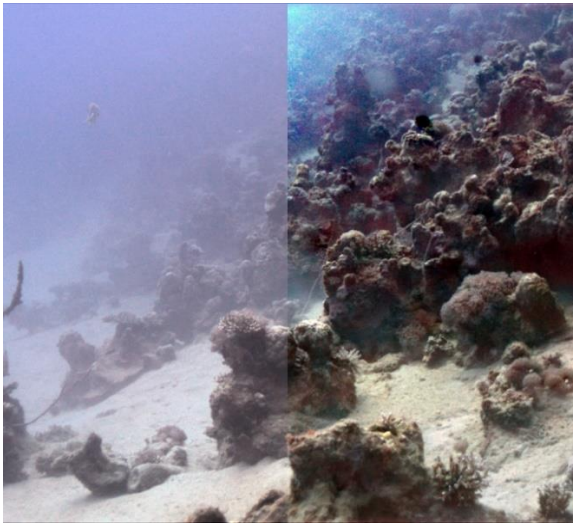

**Figure 2.** Underwater Polarization Imaging [6]. (a) Raw images taken through a polarizer; (b) The recovered image is much clearer, especially at distant objects, than a naive white-balancing attempt.

Treibitz et al. [32] (see Figure 3) proposed a fluorescence method to recover the shape of an underwater scene. Fluorescence imaging can be used to detect microorganisms in coral reefs. They also proposed a different direction lighting method to fuse the turbidity of a hazy image [33]. The fusion method removed the highlight of artificial light well. Roser et al. [34] proposed a stereo-imaging method to recover underwater images by estimating the visibility coefficients. This stereo-imaging method was designed by real-time algorithms and was applied into AUVs.

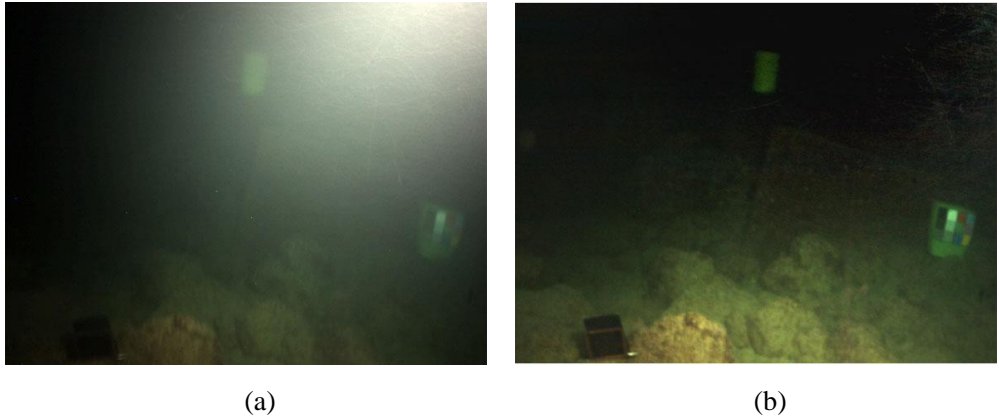
(a) (b)

**Figure 3.** Fluorescence Imaging [32]. (a) Raw image; (b) Processed result.

## 3. Underwater Image Processing
### 3.1 De-scattering

Recently, there have been many software-based approaches to underwater imaging. Depending on the outcome of the results, we can divide these approaches into two methods: wavelength compensation (sediment scattering) and color reconstruction (light absorption). To solve the scattering problem, many researchers have proposed both physical model-based methods and non-physical model-based methods. Traditional physical model-based methods are as follows. Fattal [15,16] designed a color-lines method to estimate the turbidity of haze and then used a Markov Random Field model to recover clean images. He et al. [17] proposed using a dark channel prior to estimating the depth map. Then, they employed guided filtering to refine the depth map and obtain clear images. This method can achieve real-time processing. Chiang et al. [18] proposed a wavelength compensation and dehazing method for underwater image restoration. It is the first time to consider the wavelength absorption in the imaging model. Lu et al. [19] found that some flickers exist in captured underwater images and proposed a corresponding robust ambient light estimation method and underwater median dark channel prior for de-scattering. Results of some conventional physical model-based de-scattering are shown in Figure 4.

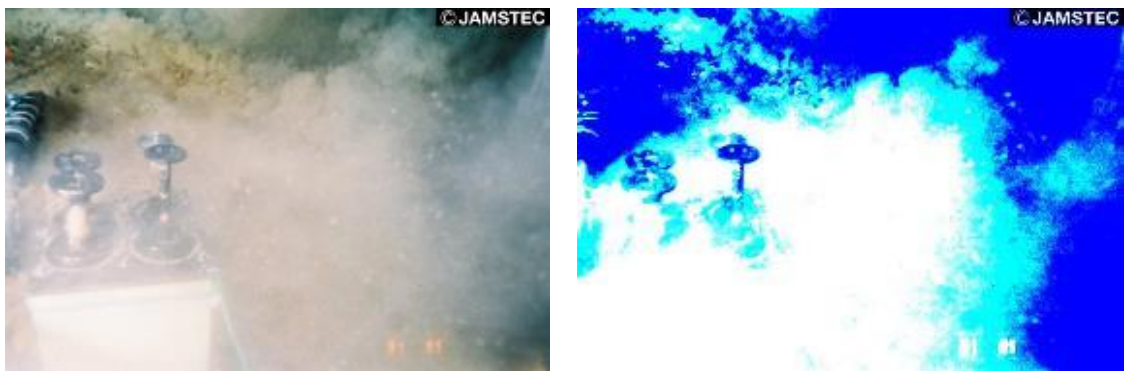

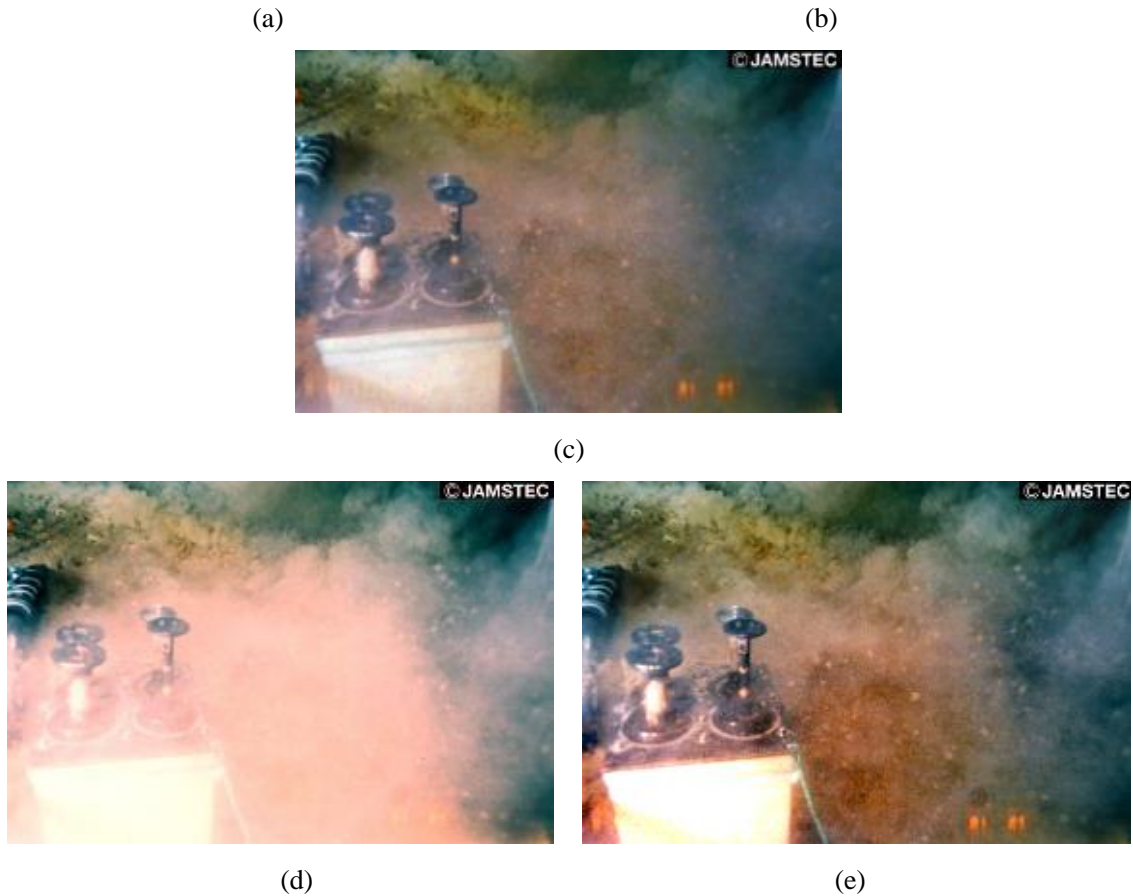

**Figure 4.** Experimental Results of Traditional Physical Model-based Methods. (a) input image; (b) Fattal's method [15]; (c) He's method [17]; (d) Chiang's method [18]; and (e) Lu's method [4].

Similarly, there have been multiple proposals for non-physical model-based methods. Garcia et al. [20] proposed local histogram equalization to address non-uniform lighting and haze. In many cases, local histogram equalization does not performs well in very dark environment. Zuiderveld et al. [21] proposed contrast limited adaptive histogram equalization (CLAHE) to adjust the target region according to an interpolation between the histograms of neighboring regions. However, non-uniform light remains on the processed image, because it operate on local regions instead of entire image. Inspired by HDR imaging, Ancuti et al. [22] proposed an espouse fusion method to combine the different exposed images via filtering. In high turbid water, the espouse fusion method cannot remove the scatter well. Galdran et al. [23] proposed a red channel-based underwater image restoration method. As mentioned in Section 1, red color channel is not always the minimum channel of the RGB color space. Gibsion et al. [24] tried to solve the scatter and noise problems simultaneously and proposed a variable Kernel size de-scattering method. After de-scattering, some halos and artifacts remain in the image. Lu et al. [25] proposed a single image dehazing method using depth map refinement. The improved bilateral filtering can smooth the depth map, while there are some residual noises exist on the image. Bazeiile et al. [26] proposed a frequency domain filtering method in YUV color space to enhance the images. The result images severe serious color distortion. Iqbal et al. [27] proposed an underwater image enhancement algorithm using an integrated color model. This method also has the color distortion issue. Experimental results of some conventional non-physical model-based methods de-scattering are shown in Figure 5.

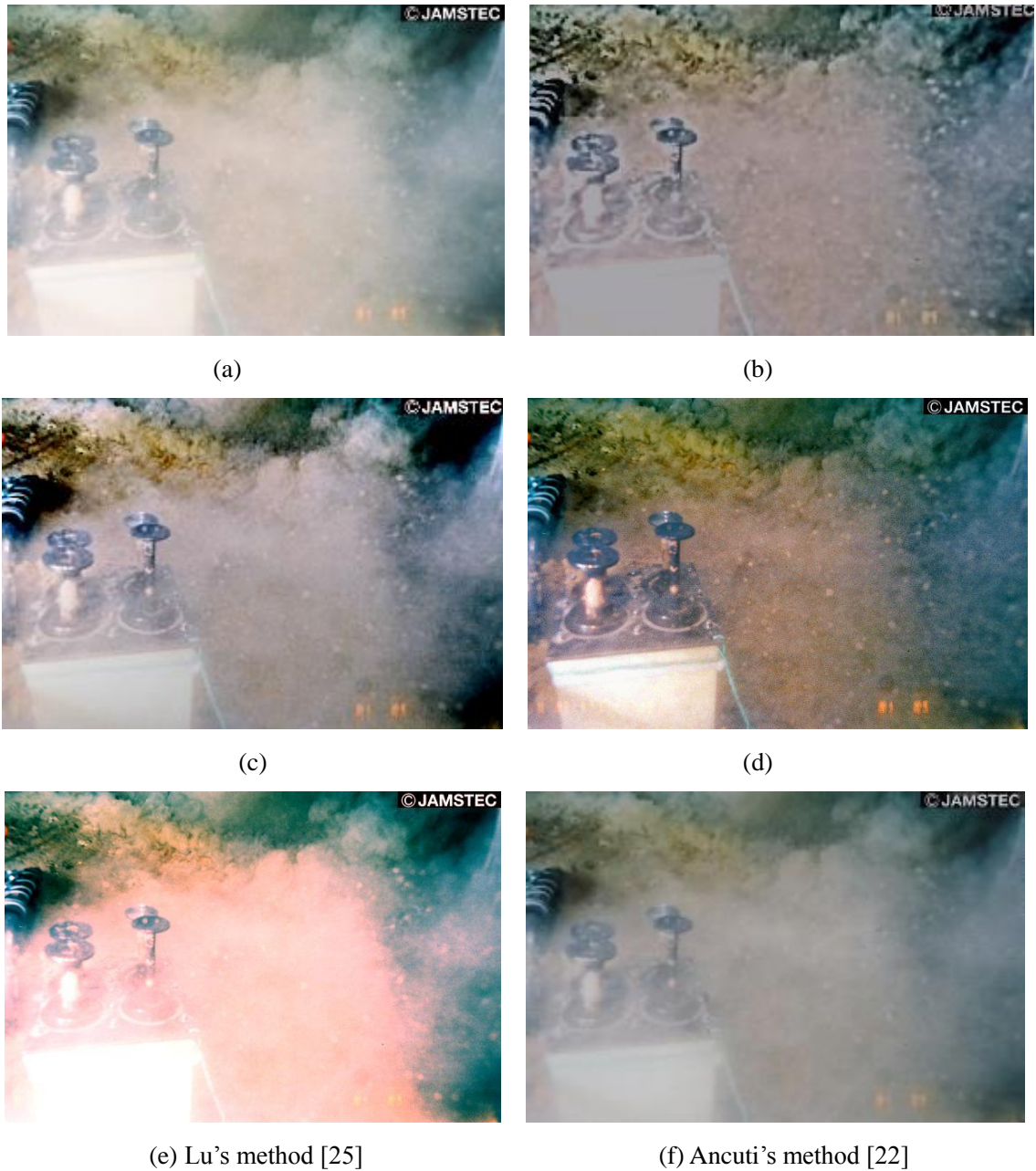

(a)                      (b)

(c)                      (d)

(e) Lu's method [25]            (f) Ancuti's method [22]

**Figure 5.** Experimental Results of Traditional Non-physical Model-based Methods. (a) Input image; (b) Bazeille's method [26]; (c) Retinex; (d) Gibson's method [24]; (e) Lu's method [25]; (f) Ancuti's method [22].

### 3.2 Underwater Image Color Restoration

As a light absorption recovering method, Torres-Mendez et al. [28] proposed a Markov Random Field (MRF) learning method to estimate the related color value of each pixel. Alhen et al. [29] developed a hyperspectral imaging and mathematical stability model to compute the attenuation coefficients using the depth map. Petit et al. [30] designed a light attenuation inversion after processing the RGB color space contraction using quaternions. Lu et al. [31] modeled the spectral response function of a camera as a function of the wavelength of the light to recover the contrast of the colors, the experimental result is shown in Figure 6. In this method, the artificial vignetting has also been solved.

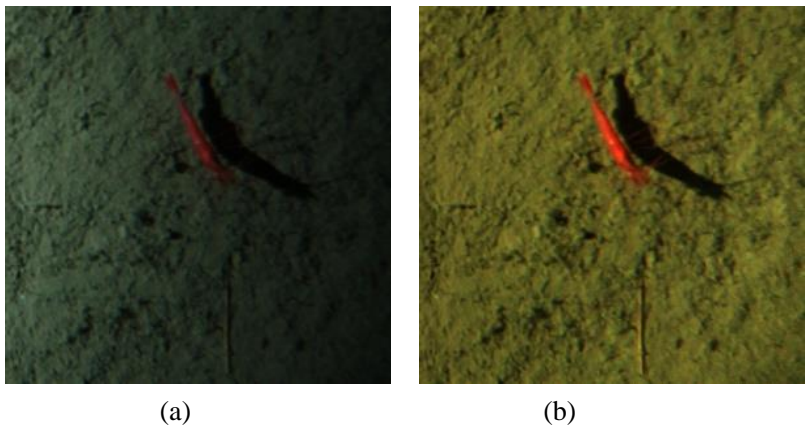

(a)                 (b)

**Figure 6.** Experimental Results of Color Restoration Methods [31]. (a) Input image; (b) Color restored image.

### 4. Underwater Image Quality Assessment

An underwater image quality assessment is also important to measure the performance of different underwater image processing methods. Wang et al. [35] proposed a structural similarity index, which treats the image degradation as a structural distortion. Panetta et al. [36] proposed a non-reference underwater image quality measure (UIQM), which combines the colorfulness measure, the sharpness measure, and the contrast measure of underwater images. Lu et al. [37] introduced a human perception method, the High-Dynamic Range Visual Difference Predictor 2, to predict both the visibility of artifacts and the overall quality in images. Yang et al. [38] proposed an underwater image quality evaluation (UCIQE), which uses a linear combination of the chroma, saturation, and contrast of underwater images in CIElab color space. Hou et al. [39] used the images' sharpness to evaluate the image quality. Arredondo et al. [40] proposed a mean angular error to assess the robustness and behavior of methods with respect to underwater noises. Lu et al. [41] proposed a new index, $Q_u$, which can evaluate the similarity of both the structures and colors of underwater images.

There are also existed some other issues for underwater optical image processing. The non-uniform artificial lighting, inhomogeneous de-scattering, high turbidity image reconstruction, image reflection, and computational underwater imaging et al. will be focused in near future. Other related research fields will be studied for underwater image processing, such as deep learning [44], cloud computing [45, 46], and internet of things [47].

In Table I, we summarize the different underwater image enhancement methods along with some representative studies.

**Table I.** Categorization of the underwater image restoration methods and their representative studies.

| Algorithm | Model's Characteristics | | Method |
|---|---|---|---|
| Review | - | | Kocak et al. [8] |
| | - | | Lu et al. [2] |
| | - | | Schettini et al. [42] |
| | - | | Hou [43] |
| Hardware-based Methods | Polarization | | Yemelyanov et al. [5], Two channel polarization |
| | | | Schechner et al. [6], Polarization filtering |
| | Range-gated imaging | | Tan et al. [7,9], Range-gated underwater laser imaging |
| | | | Li et al. [10], Speckle reduction of range-gated imaging |
| | Fluorescence imaging | | Treibitz et al. [32], Multi-lighting |
| | | | Treibitz et al. [33], Multi-images fusion |
| | Stereo imaging | | Roser et al. [34], Visibility coefficients estimation |
| Software-based Methods | Wavelength compensation | Physical Model | Fattal [15], PCA dehazing |
| | | | Fattal [16], Color-lines dehazing |
| | | | He et al. [17], DCP |
| | | | Chiang et al. [18], WCID |
| | | | Lu et al. [19], De-flickering De-scattering |
| | | Non-physical model | Garcia et al. [20], Local histogram equalization |
| | | | Zuiderveld et al. [21], Contrast |

| | | | limited adaptive histogram equalization |
|---|---|---|---|
| | | | Ancuti et al. [22], HDR fusion |
| | | | Galdran et al. [23], Red channel dehazing |
| | | | Gibsion et al. [24], Wiener filtering |
| | | | Lu et al. [25], Single image dehazing |
| | | | Bazeiile et al. [26], Pre-filtering |
| | | | Iqbal et al. [27], Integrated color model |
| | | | Arnold-Bos et al. [11], Pre-processing |
| | | | Rizzi et al. [12], Unsupervised global and local color correction |
| | | | Arnold-Bos et al. [13], Model-free denoising |
| | | | Chambah et al. [14], Color constancy |
| | | Color reconstruction | Torres-Mendez et al. [28], Markov Random Field learning |
| | | | Alhen et al. [29], Hyperspectral imaging |
| | | | Petit et al. [30], Light attenuation inversion |
| | | | Lu et al. [31], Spectral response function |
| Image Quality Assessment | | Reference | Wang et al. [35], SSIM |
| | | | Lu et al. [41], $Q_u$ |
| | | | Arredondo et al. [40], Mean angular error |
| | | Non-reference | Panetta et al. [36], UIQM |
| | | | Lu et al. [37], QMOS |
| | | | Yang et al. [38], UCIQUE |
| | | | Hou et al. [39], WGSA metric |

## 5. Conclusions and Future Trends

In this paper, we presented a comprehensive review of underwater image processing. We divided the underwater image processing methods into two categories according to their imaging types. The state-of-the-art approaches of the two classes were discussed and analyzed in detail. For software-based underwater image processing, wavelength compensation approach, e.g. physical model, non-physical model and color reconstruction approach are discussed. Finally, the quality assessment methods and future trends are summarized.

**Acknowledgements**

This work was supported by JSPS KAKENHI (15F15077), Leading Initiative for Excellent Young Researcher (LEADER) of Ministry of Education, Culture, Sports, Science and Technology-Japan (16809746), Research Fund of Chinese Academy of Sciences (MGE2015KG02), Research Fund of State Key Laboratory of Marine Geology in Tongji University (MGK1608), and Research Fund of State Key Laboratory of Ocean Engineering in Shanghai Jiaotong University (1315; 1510).